\documentclass[runningheads]{llncs}\usepackage[]{graphicx}\usepackage[]{color}
\makeatletter
\def\maxwidth{ %
  \ifdim\Gin@nat@width>\linewidth
    \linewidth
  \else
    \Gin@nat@width
  \fi
}
\makeatother

\definecolor{fgcolor}{rgb}{0.345, 0.345, 0.345}

\usepackage{framed}
\makeatletter
 {\par\unskip\endMakeFramed%
 \at@end@of@kframe}
\makeatother

\definecolor{shadecolor}{rgb}{.97, .97, .97}
\definecolor{messagecolor}{rgb}{0, 0, 0}
\definecolor{warningcolor}{rgb}{1, 0, 1}
\definecolor{errorcolor}{rgb}{1, 0, 0}
\usepackage{alltt}
\usepackage[utf8]{inputenc}
\usepackage{graphicx}
\usepackage{enumerate}
\usepackage{cite}
\usepackage{amsmath}
\usepackage{amssymb}
\usepackage{mathabx}
\usepackage{bbold}
\usepackage{float}
\usepackage[ruled,vlined,linesnumbered, boxed, english]{algorithm2e}
\usepackage[misc]{ifsym}
\usepackage{hyperref}

\IfFileExists{upquote.sty}{\usepackage{upquote}}{}
\begin{document}

\title{Sampling, Intervention, Prediction, Aggregation: A Generalized Framework for Model-Agnostic Interpretations}
\titlerunning{A Generalized Framework for Model-Agnostic Interpretations}

\author{Christian~A.~Scholbeck (\Letter), Christoph Molnar, Christian Heumann, Bernd Bischl, Giuseppe Casalicchio}
\authorrunning{CA Scholbeck et al.}

\institute{Department of Statistics, Ludwig-Maximilians-University Munich, \\
  Ludwigstr. 33, 80539 Munich, Germany \\
  \email{christian.scholbeck@stat.uni-muenchen.de}%\\
}
\maketitle

\begin{abstract}

\noindent Model-agnostic interpretation techniques allow us to explain the behavior of any predictive model. Due to different notations and terminology, it is difficult to see how they are related. A  unified  view  on  these  methods  has been missing. We present the generalized SIPA (sampling, intervention, prediction, aggregation) framework of work stages for model-agnostic interpretations and demonstrate how several prominent methods for feature effects can be embedded into the proposed framework. Furthermore, we extend the framework to feature importance computations by pointing out how variance-based and performance-based importance measures are based on the same work stages. The SIPA framework reduces the diverse set of model-agnostic techniques to a single methodology and establishes a common terminology to discuss them in future work.

\keywords{Interpretable Machine Learning $\vert$ Explainable AI $\vert$ Feature Effect $\vert$ Feature Importance $\vert$ Model-Agnostic $\vert$ Partial Dependence}

\end{abstract}

\section{Introduction and Related Work}

There has been an ongoing debate about the lacking interpretability of machine learning (ML) models. As a result, researchers have put in great efforts developing techniques to create insights into the workings of predictive black box models. Interpretable machine learning \cite{molnar_iml} serves as an umbrella term for all interpretation methods in ML. We make the following distinctions:

\begin{enumerate}[(i)]
\item
\textit{Feature effects or feature importance}: Feature effects indicate the direction and magnitude of change in predicted outcome due to changes in feature values. Prominent methods include the individual conditional expectation (ICE) \cite{goldstein_ice} and partial dependence (PD) \cite{friedman_pdp}, accumulated local effects (ALE) \cite{apley_ale}, Shapley values \cite{strumbelj_shapley} and local interpretable model-agnostic explanations (LIME) \cite{ribeiro_lime}. The feature importance measures the importance of a feature to the model behavior. This includes variance-based measures like the feature importance ranking measure (FIRM) \cite{greenwell_varimp}, \cite{zien_firm} and performance-based measures like the permutation feature importance (PFI) \cite{rudin_modelreliance}, individual conditional importance (ICI) and partial importance (PI) curves \cite{casalicchio_featureimp}, as well as the Shapley feature importance (SFIMP) \cite{casalicchio_featureimp}. Input gradients were proposed by \cite{hechtlinger_inputgradient} as a model-agnostic tool for both effects and importance that essentially equals marginal effects (ME) \cite{leeper_margins}, which have a long tradition in statistics. They also define an average input gradient which corresponds to the average marginal effect (AME).
\item
\textit{Intrinsic or post-hoc interpretability}: Linear models (LM), generalized linear models (GLM), classification and regression trees (CART) or rule lists \cite{rudin_rulelists} are examples for intrinsically interpretable models, while random forests (RF), support vector machines (SVM), neural networks (NN) or gradient boosting (GB) models can only be interpreted post-hoc. Here, the interpretation process is detached from and takes place after the model fitting process, e.g., with the ICE, PD or ALEs.
\item
\textit{Model-specific or model-agnostic interpretations}: Interpreting model coefficients of GLMs or deriving a decision rule from a classification tree is a model-specific interpretation. Model-agnostic methods such as the ICE, PD or ALEs can be applied to any model.
\item
\textit{Local or global explanations}: Local explanations like the ICE evaluate the model behavior when predicting for one specific observation. Global explanations like the PD interpret the model for the entire input space. Furthermore, it is possible to explain model predictions for a group of observations, e.g., on intervals. In a lot of cases, local and global explanations can be transformed into one another via (dis-)aggregation, e.g., the ICE and PD.
\end{enumerate}

\par
\textit{Motivation}: Research in model-agnostic interpretation methods is complicated by the variety of different notations and terminology. It turns out that deconstructing model-agnostic techniques into sequential work stages reveals striking similarities. In \cite{lundberg_shap} the authors propose a unified framework for model-agnostic interpretations called SHapley Additive exPlanations (SHAP). However, the SHAP framework only considers Shapley values or variations thereof (KernelSHAP and TreeSHAP). The motivation for this research paper is to provide a more extensive survey on model-agnostic interpretation methods, to reveal similarities in their computation and to establish a framework with common terminology that is applicable to all model-agnostic techniques.
\par
\textit{Contributions:} In Section \hyperref[sec:framework_effects]{4} we present the generalized SIPA (sampling, intervention, prediction, aggregation) framework of work stages for model-agnostic techniques. We proceed to demonstrate how several methods to estimate feature effects (MEs, ICE and PD, ALEs, Shapley values and LIME) can be embedded into the proposed framework. Furthermore, in Section \hyperref[sec:feature_importance]{5} and \hyperref[sec:framework_importance]{6} we extend the framework to feature importance computations by pointing out how variance-based (FIRM) and performance-based (ICI and PI, PFI and SFIMP) importance measures are based on the same work stages. By using a unified notation, we also reveal how the methods are related.

\section{Notation and Preliminaries}

Consider a p-dimensional feature space $\mathcal{X}_P = \mathcal{X}_1 \times \dots \times \mathcal{X}_p$ with the feature index set $P = \{1, \dots, p\}$ and a target space $\mathcal{Y}$. We assume an unknown functional relationship $f$ between $\mathcal{X}_P$ and $\mathcal{Y}$. A supervised learning model $\hat{f}$ attempts to learn this relationship from an i.i.d. training sample that was drawn from the unknown probability distribution $\mathcal{F}$ with the sample space $\mathcal{X}_P \times \mathcal{Y}$. The random variables generated from the feature space are denoted by $X = (X_1, \dots, X_p)$. The random variable generated from the target space is denoted by $Y$. We draw an i.i.d. sample of test data $\mathcal{D}$ with $n$ observations from $\mathcal{F}$. The vector $x^{(i)} = (x_1^{(i)}, \dots, x_p^{(i)}) \in \mathcal{X}_P$ corresponds to the feature values of the $i$-th observation that are associated with the observed target value $y^{(i)} \in \mathcal{Y}$. The vector $x_j = (x_j^{(1)}, \dots, x_j^{(n)})^\top$ represents the realizations of $X_j$. The generalization error $GE(\hat{f}, \mathcal{F})$ corresponds to the expectation of the loss function $\mathcal{L}$ on unseen test data from $\mathcal{F}$ and is estimated by the average loss on $\mathcal{D}$.
\begin{align*}
GE(\hat{f}, \mathcal{F}) &= \mathbb{E}\left[ \mathcal{L}(\hat{f}(X_1, \dots, X_p), Y) \right] \\
\widehat{GE}(\hat{f}, \mathcal{D}) &= \frac{1}{n} \sum_{i = 1}^n \mathcal{L}(\hat{f}(x_1^{(i)}, \dots, x_p^{(i)}), y^{(i)})
\end{align*}
\par
A variety of model-agnostic techniques is used to interpret the prediction function $\hat{f}(x_1, \dots, x_p)$ with the sample of test data $\mathcal{D}$. We estimate the effects and importance of a subset of features with index set $S$ ($S\subseteq P$). A vector of feature values $x \in \mathcal{X}_P$ can be partitioned into two vectors $x_S$ and $x_{\setminus S}$ so that $x = (x_S, x_{\setminus S})$. The corresponding random variables are denoted by $X_S$ and $X_{\setminus S}$. Given a model-agnostic technique where $S$ only contains a single element, the corresponding notations are $X_j, X_{\setminus j}$ and $x_{j}, x_{\setminus j}$.
\par
The partial derivative of the trained model $\hat{f}(x_j, x_{\setminus j})$ with respect to $x_j$ is numerically approximated with a symmetric difference quotient \cite{leeper_margins}.
\begin{equation*}
\lim \limits_{h \to 0} \frac{\hat{f}(x_j + h, x_{\setminus j}) - \hat{f}(x_j, x_{\setminus j})}{h} \approx \frac{\hat{f}(x_j + h, x_{\setminus j}) - \hat{f}(x_j - h, x_{\setminus j})}{2h}, \quad h > 0
\end{equation*}

\noindent A term of the form $\hat{f}(x_j + h, x_{\setminus j}) - \hat{f}(x_j - h, x_{\setminus j})$ is called a finite difference (FD) of predictions with respect to $x_j$.

\begingroup
\allowdisplaybreaks
\begin{equation*}
{FD}_{\hat{f}, j}(x_j, x_{\setminus j}) = \hat{f}(x_j + h, x_{\setminus j}) - \hat{f}(x_j - h, x_{\setminus j})
\end{equation*}
\endgroup

\section{Feature Effects}
\label{sec:feature_effects}

\textit{Partial dependence (PD) and individual conditional expectation (ICE)}: First suggested by \cite{friedman_pdp}, the PD is defined as the dependence of the prediction function on $x_S$ after all remaining features $X_{\setminus S}$ have been marginalized out \cite{goldstein_ice}. The PD is estimated via Monte Carlo integration.

\begin{align}
{PD}_{\hat{f}, S}(x_S) &= \mathbb{E}_{X_{\setminus S}}\left[\hat{f}(x_S, X_{\setminus S})\right] = \int \hat{f}(x_S, X_{\setminus S}) \; d\mathcal{P}(X_{\setminus S})
\label{eq:pd} \\
\widehat{PD}_{\hat{f}, S}(x_S) &= \frac{1}{n} \sum\limits_{i = 1}^n \hat{f}(x_S, x_{\setminus S}^{(i)}) \nonumber
\end{align}

\noindent The PD is a useful feature effect measure when features are not interacting \cite{friedman_pdp}. Otherwise it can obfuscate the relationships in the data \cite{casalicchio_featureimp}. In that case, the individual conditional expectation (ICE) can be used instead \cite{goldstein_ice}. The $i$-th ICE corresponds to the expected value of the target for the $i$-th observation as a function of $x_S$, conditional on $x_{\setminus S}^{(i)}$.
\begin{align*}
\widehat{ICE}_{\hat{f}, S}^{(i)}(x_S) &= \hat{f}(x_S, x_{\setminus S}^{(i)})
\end{align*}

\noindent The ICE disaggregates the global effect estimates of the PD to local effect estimates for single observations. Given $\vert S \vert = 1$, the ICE and PD are also referred to as ICE and PD curves. The ICE and PD suffer from extrapolation when features are correlated, because the permutations used to predict are located in regions without any training data \cite{apley_ale}.
\par
\textit{Accumulated local effects (ALE)}: In \cite{apley_ale} ALEs are presented as a feature effect measure for correlated features that does not extrapolate. The idea of ALEs is to take the integral with respect to $X_j$ of the first derivative of the prediction function with respect to $X_j$. This creates an accumulated partial effect of $X_j$ on the target variable while simultaneously removing additively linked effects of other features. The main advantage of not extrapolating stems from integrating with respect to the conditional distribution of $X_{\setminus j}$ on $X_j$ instead of the marginal distribution of $X_{\setminus j}$ \cite{apley_ale}. Let $z_{0, j}$ denote the minimum value of $x_j$. The first order ALE of the $j$-th feature at point $x$ is defined as:

\begingroup
\allowdisplaybreaks
\begin{align}
{ALE}_{\hat{f}, j}(x) &= \int_{z_{0, j}}^{x} \mathbb{E}_{X_{\setminus j} \vert X_j} \left[\frac{\partial \hat{f}(X_j, X_{\setminus j})}{\partial X_j} \bigg \vert X_j = z_j \right] dz_j - constant \nonumber \\
&= \int_{z_{0, j}}^{x} \left[ \int \frac{\partial \hat{f}(z_j, X_{\setminus j})}{\partial z_j} \; d\mathcal{P}(X_{\setminus j} | z_j) \right] dz_j - constant \label{eq:ale_estimation}
\end{align}
\endgroup

\noindent A constant is subtracted in order to center the plot. We estimate the first order ALE in three steps. First, we divide the value range of $x_j$ into a set of intervals and compute a finite difference (FD) for each observation. For each $i$-th observation, $x_j^{(i)}$ is substituted by the corresponding right and left interval boundaries. Then the predictions with both substituted values are subtracted in order to receive an observation-wise FD. Second, we estimate local effects by averaging the FDs inside each interval. This replaces the inner integral in Eq. (\ref{eq:ale_estimation}). Third, the accumulation of all local effects up to the point of interest replaces the outer integral in Eq. (\ref{eq:ale_estimation}), i.e., the interval-wise average FDs are summed up.
\par
The second order ALE is the bivariate extension of the first order ALE. It is important to note that first order effect estimates are subtracted from the second order estimates. In \cite{apley_ale} the authors further lay out the computations necessary for higher order ALEs.
\par
\textit{Marginal effects (ME)}: MEs are an established technique in statistics and often used to interpret non-linear functions of coefficients in GLMs like logistic regression. The ME corresponds to the first derivative of the prediction function with respect to a feature at specified values of the input space. It is estimated by computing an observation-wise FD. The average marginal effect (AME) is the average of all MEs that were estimated with observed feature values \cite{bartus_marginal_effects}. Although there is extensive literature on MEs, this concept was suggested by \cite{hechtlinger_inputgradient} as a novel method for ML and referred to as the input gradient. Derivatives are also often utilized as a feature importance metric.
\par
\textit{Shapley value}: Originating in coalitional game theory \cite{strumbelj_shapley}, the Shapley value is a local feature effect measure that is based on a set of desirable axioms. In coalitional games, a set of $p$ players, denoted by $P$, play games and join coalitions. They are rewarded with a payout. The characteristic function $v: 2^{p} \rightarrow \mathbb{R}$ maps all player coalitions to their respective payouts \cite{casalicchio_featureimp}. The Shapley value is a player's average contribution to the payout, i.e., the marginal increase in payout for the coalition of players, averaged over all possible coalitions. For Shapley values as feature effects, predicting the target for a single observation corresponds to the game and a coalition of features represents the players. Shapley regression values were first developed for linear models with multicollinear features \cite{lipovetsky_shapley}. A model-agnostic Shapley value was first introduced in \cite{strumbelj_shapley}.
\par
Consider the expected prediction for a single vector of feature values $x$, conditional on only knowing the values of features with indices in $K$ ($K \subseteq P$), i.e., the features $X_{\setminus K}$ are marginalized out. This essentially equals a point (or a line, surface etc. depending on the power of $K$) on the PD from Eq. (\ref{eq:pd}).
\begingroup
\allowdisplaybreaks
\begin{equation}
% &\phantom{{}={}} \mathbb{E}\left[\hat{f} \; \Big \vert \; X_k = x_k, \forall k \in K \right] \nonumber \\
\mathbb{E}_{X_{\setminus K}}\left[\hat{f}(x_K, X_{\setminus K})\right] = \int \hat{f}(x_K, X_{\setminus K}) \; d\mathcal{P} (X_{\setminus K}) = \widehat{PD}_{\hat{f}, K}(x_K)
\label{eq:shapley_pd}
\end{equation}
\endgroup

\noindent Eq. (\ref{eq:shapley_pd}) is shifted by the mean prediction and used as a payout function $v_{PD}(x_K)$, so that an empty set of features ($K = \emptyset$) results in a payout of zero \cite{casalicchio_featureimp}.
\begin{align*}
v_{PD}(x_K) &= \mathbb{E}_{X_{\setminus K}}\left[ \hat{f}(x_K, X_{\setminus K}) \right] - \mathbb{E}_{X_{K \cup (P \setminus K)}}\left[ \hat{f}(X_K, X_{\setminus K}) \right] \\
&= \widehat{PD}_{\hat{f}, K}(x_K) - \widehat{PD}_{\hat{f}, \emptyset}(x_{\emptyset}) \\
&= \widehat{PD}_{\hat{f}, K}(x_K) - \frac{1}{n}\sum_{i = 1}^n \hat{f}(x_K^{(i)}, x_{\setminus K}^{(i)}) 
\end{align*}
The marginal contribution $\Delta_j(x_K) $ of a feature value $x_j$ joining the coalition of feature values $x_K$ is:
\begin{equation*}
\Delta_j(x_K) = v_{PD}(x_{K \cup \{j\}}) - v_{PD}(x_K) = \widehat{PD}_{\hat{f}, K \cup \{j\}}(x_{K \cup \{j\}}) - \widehat{PD}_{\hat{f}, K}(x_K)
\end{equation*}
The exact Shapley value of the $j$-th feature for a single vector of feature values $x$ corresponds to:

% \begingroup
% \allowdisplaybreaks
\begin{align*}
\widehat{Shapley}_{\hat{f}, j} &= \sum_{K \, \subseteq \, P \setminus \{j\}} \frac{|K|!(|P| - |K| - 1)!}{|P|!} \; \Delta_j(x_K) \\
&= \sum_{K \, \subseteq \, P \setminus \{j\}} \frac{|K|!(|P| - |K| - 1)!}{|P|!}\left[ \widehat{PD}_{\hat{f}, K \cup \{j\}}(x_{K \cup \{j\}}) - \widehat{PD}_{\hat{f}, K}(x_K) \right]
\end{align*}
% \endgroup

Shapley values are computationally expensive because the PD function has a complexity of $\mathcal{O}(N^2)$. Computations can be sped up by Monte Carlo sampling \cite{strumbelj_shapley}.
Furthermore, in \cite{lundberg_shap} the authors propose a distinct variant to compute Shapley values called SHapley Additive exPlanations (SHAP).
\par
\textit{Local interpretable model-agnostic explanations (LIME)}: In contrast to all previous techniques which are based on interpreting a single model, LIME \cite{ribeiro_lime} locally approximates the black box model with an intrinsically interpretable surrogate model. Given a single vector of feature values $x$, we first perturb $x_j$ around a sufficiently close neighborhood while $x_{\setminus j}$ is kept constant. Then we predict with the perturbed feature values. The predictions are weighted by the proximity of the corresponding perturbed values to the original feature value. Finally, an intrinsically interpretable model is trained on the weighted predictions and interpreted instead.

\section{Generalized Framework}
\label{sec:framework_effects}

Although the techniques presented in Section \ref{sec:feature_effects} are seemingly unrelated, they all work according to the exact same principle. Instead of trying to inspect the inner workings of a non-linear black box model, we evaluate its predictions when changing inputs. We can deconstruct model-agnostic techniques into a framework of four work stages: sampling, intervention, prediction, aggregation (SIPA). The software package \texttt{iml} \cite{molnar_imlpackage} was inspired by the SIPA framework.
\par
We first sample a subset (\textbf{sampling stage}) to reduce computational costs, e.g., we select a random set of available observations to evaluate as ICEs. In order to change the predictions made by the black box model, the data has to be manipulated. Feature values can be set to values from the observed marginal distributions (ICEs and PD or Shapley values), or to unobserved values (FD based methods such as MEs and ALEs). This crucial step is called the \textbf{intervention stage}. During the \textbf{prediction stage}, we predict on previously intervened data. This requires an already trained model, which is why model-agnostic techniques are always post-hoc. The predictions are further aggregated during the \textbf{aggregation stage}. Often, the predictions resulting from the prediction stage are local effect estimates, and the ones resulting from the aggregation stage are global effect estimates.
\par
\begin{figure}
\centering
 \includegraphics[width = \textwidth]{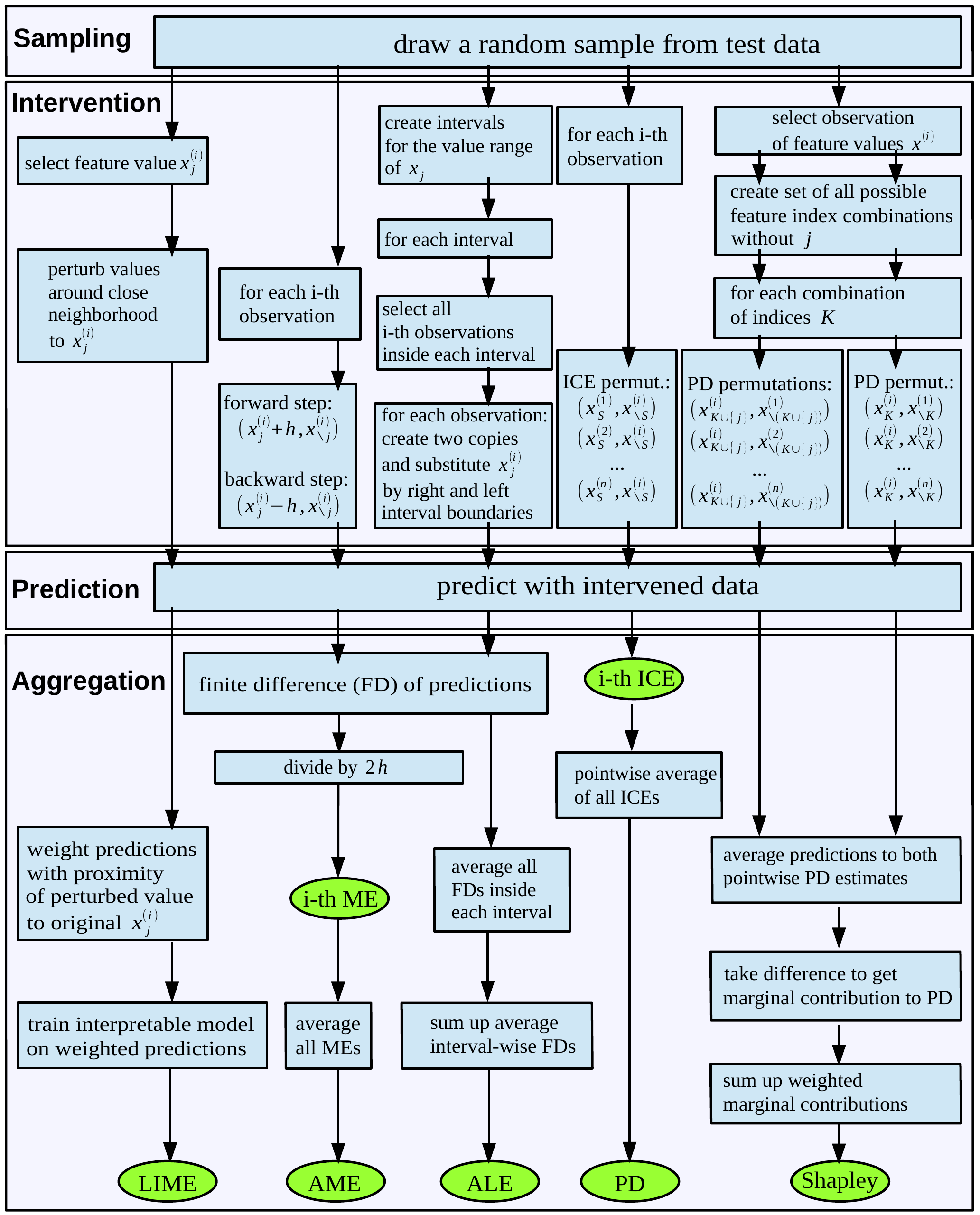}
 \caption{We demonstrate how all presented model-agnostic methods for feature effects are based on the SIPA framework. For every method, we assign each computational step to the corresponding generalized SIPA work stage. Contrary to all other methods, LIME is based on training an intrinsically interpretable model during the aggregation stage. We consider training a model to be an aggregation, because it corresponds to an optimization problem where the training data is aggregated to a function. For reasons of simplicity, we do not differentiate between the actual functions or values and their estimates. \label{fig:framework_embeddings}}
\end{figure}

In Fig. \ref{fig:framework_embeddings}, we demonstrate how all presented techniques for feature effects are based on the SIPA framework. Although LIME is a special case as it is based on training a local surrogate model, we argue that it is also based on the SIPA framework as training a surrogate model can be considered an aggregation of the training data to a function. 

\section{Feature Importance}
\label{sec:feature_importance}

We categorize model-agnostic importance measures into two groups: variance-based and performance-based.
\par
\textit{Variance-based}: A mostly flat trajectory of a single ICE curve implies that in the underlying predictive model, varying $x_j$ does not affect the prediction for this specific observation. If all ICE curves are shaped similarly, the PD can be used instead. In \cite{greenwell_varimp} the authors propose a measure for the curvature of the PD as a feature importance metric. Let the average value of the estimated PD of the $j$-th feature be denoted by $\overline{\widehat{PD}}_{\hat{f}, j}(x_j) =  \frac{1}{n} \sum_{i = 1}^{n} \widehat{PD}_{\hat{f}, j}(x_j^{(i)})$. The estimated importance $\widehat{\text{IMP}}_{\widehat{PD}, j}$ of the $j$-th feature corresponds to the standard deviation of the feature's estimated PD function. The flatter the PD, the smaller its standard deviation and therefore the importance metric. For categorial features, the range of the PD is divided by 4. This is supposed to represent an approximation to the estimate of the standard deviation for small to medium sized samples \cite{greenwell_varimp}.

\begin{align} \label{eq:imp_pd}
\widehat{\text{IMP}}_{\widehat{PD}, j} =
\begin{cases}
    \sqrt{ \frac{1}{n-1} \, \sum\limits_{i = 1}^{n} \left [\widehat{PD}_{\hat{f}, j}(x_j^{(i)}) - \overline{\widehat{PD}}_{\hat{f}, j}(x_j) \right]^2} & \text{$x_j$ continuous} \\
    \frac{1}{4} \; \left[ max \left\{\widehat{PD}_{\hat{f}, j}(x_j) \right\} - min \left\{\widehat{PD}_{\hat{f}, j}(x_j) \right\} \right] & \text{$x_j$ categorial}
\end{cases}
\end{align}

In \cite{zien_firm} the authors propose the feature importance ranking measure (FIRM). They define a conditional expected score (CES) function for the $j$-th feature.

\begin{equation}
CES_{\hat{f}, j}(v) = \mathbb{E}_{X_{\setminus j}}\left[\hat{f}(x_j, X_{\setminus j}) \; \big \vert \; x_j = v \right]
\label{eq:firm_ces}
\end{equation}

\noindent It turns out that Eq. (\ref{eq:firm_ces}) is equivalent to the PD from Eq. (\ref{eq:pd}), conditional on $x_j = v$.

\begin{align*}
CES_{\hat{f}, j}(v) &= \mathbb{E}_{X_{\setminus j}}\left[\hat{f}(v, X_{\setminus j})\right] \\
&= PD_{\hat{f}, j}(v)
\end{align*}

\noindent The FIRM corresponds to the standard deviation of the CES function with all values of $x_j$ used as conditional values. This in turn is equivalent to the standard deviation of the PD. The FIRM is therefore equivalent to the feature importance metric in Eq. (\ref{eq:imp_pd}).

\begin{equation*}
\widehat{FIRM}_{\hat{f}, j} = \sqrt{Var(\widehat{CES}_{\hat{f}, j}(x_j))} = \sqrt{Var(\widehat{PD}_{\hat{f}, j}(x_j))} = \widehat{\text{IMP}}_{\widehat{PD}, j}
\label{eq:firm_importance}
\end{equation*}

\par
\textit{Performance-based}: The permutation feature importance (PFI), originally developed by \cite{breiman_randomforests} as a model-specific tool for random forests, was described as a model-agnostic one by \cite{fisher_pfi}. If feature values are shuffled in isolation, the relationship between the feature and the target is broken up. If the feature is important for the predictive performance, the shuffling should result in an increased loss \cite{casalicchio_featureimp}. Permuting $x_j$ corresponds to drawing from a new random variable $\tilde{X}_j$ that is distributed like $X_j$ but independent of $X_{\setminus j}$ \cite{casalicchio_featureimp}. The model-agnostic PFI measures the difference between the generalization error (GE) on data with permuted and non-permuted values. 
\begin{equation*}
PFI_{\hat{f}, j} = \mathbb{E} \left[\mathcal{L}(\hat{f}(\tilde{X}_j, X_{\setminus j}), Y) \right] - \mathbb{E} \left[\mathcal{L}(\hat{f}(X_j, X_{\setminus j}), Y) \right]
\end{equation*}
Let the permutation of $x_j$ be denoted by $\tilde{x}_j$. Consider the sample of test data $\mathcal{D}_j$ where $x_j$ has been permuted, and the non-permuted sample $\mathcal{D}$. The PFI estimate is given by the difference between GE estimates with permuted and non-permuted values.
\begin{align}
% \widehat{PFI}_S = \frac{1}{n} \sum_{i = 1}^N \mathcal{L}(\hat{f}(\tilde{x}_S^{(i)}, x_C^{(i)}), Y) - \frac{1}{n} \sum_{i = 1}^N \mathcal{L}(\hat{f}(x_S^{(i)}, x_C^{(i)}), Y)
\widehat{PFI}_{\hat{f}, j} &= \widehat{GE}(\hat{f}, \mathcal{D}_j) - \widehat{GE}(\hat{f}, \mathcal{D}) \nonumber \\
&= \frac{1}{n} \sum_{i = 1}^n \mathcal{L}(\hat{f}(\tilde{x}_j^{(i)}, x_{\setminus j}^{(i)}), y^{(i)}) - \frac{1}{n} \sum_{i = 1}^n \mathcal{L}(\hat{f}(x_j^{(i)}, x_{\setminus j}^{(i)}), y^{(i)}) 
\label{eq:pfi}
\end{align}

\par
In \cite{casalicchio_featureimp} the authors propose individual conditional importance (ICI) and partial importance (PI) curves as visualization techniques that disaggregate the global PFI estimate. They are based on the same principle as the ICE and PD. The ICI visualizes the influence of a feature on the predictive performance for a single observation, while the PI visualizes the average influence of a feature for all observations. Consider the prediction for the $i$-th observation with observed values $\hat{f}(x_j^{(i)}, x_{\setminus j}^{(i)})$ and the prediction $\hat{f}(x_j^{(l)}, x_{\setminus j}^{(i)})$ where $x_j^{(i)}$ was replaced by a value $x_j^{(l)}$ from the marginal distribution of observed values $x_j$. The change in loss is given by:

\begin{equation*}
\Delta \mathcal{L}^{(i)}(x_j^{(l)}) = \mathcal{L}(\hat{f}(x_j^{(l)}, x_{\setminus j}^{(i)})) - \mathcal{L}(\hat{f}(x_j^{(i)}, x_{\setminus j}^{(i)}))
\end{equation*}

\noindent The ICI curve of the $i$-th observation plots the value pairs $(x_j^{(l)}, \Delta \mathcal{L}^{(i)}(x_j^{(l)}))$ for all $l$ values of $x_j$. The PI curve is the pointwise average of all ICI curves at all $l$ values of $x_j$. It plots the value pairs $(x_j^{(l)}, \frac{1}{n} \sum_{i = 1}^n \Delta \mathcal{L}^{(i)}(x_j^{(l)}))$ for all $l$ values of $x_j$. Substituting values of $x_j$ essentially resembles shuffling them. The authors demonstrate how averaging the values of the PI curve results in an estimation of the global PFI.
\begin{equation*}
% \widehat{PFI_S} = \frac{1}{n} \sum_{i = 1}^n \widehat{PFI_S}^{(i)} = \frac{1}{n} \sum_{i = 1}^n \widehat{PI}_S(x_S^{(i)})
% \widehat{PFI}_{\hat{f}, j} = \frac{1}{n} \sum_{i = 1}^n \widehat{PFI}_{\hat{f}, j}^{(i)} = \frac{1}{n}\sum_{l = 1}^n \frac{1}{n} \sum_{i = 1}^n \Delta \mathcal{L}^{(i)}(x_S^{(l)})
\widehat{PFI}_{\hat{f}, j} = \frac{1}{n}\sum_{l = 1}^n \frac{1}{n} \sum_{i = 1}^n \Delta \mathcal{L}^{(i)}(x_j^{(l)})
\end{equation*}

\par
Furthermore, a feature importance measure called Shapley feature importance (SFIMP) was proposed in \cite{casalicchio_featureimp}. Shapley importance values based on model refits with distinct sets of features were first introduced by \cite{cohen_shapley} for feature selection. This changes the behavior of the learning algorithm and is not helpful to evaluate a single model, as noted by \cite{casalicchio_featureimp}. The SFIMP is based on the same computations as the Shapley value but replaces the payout function with one that is sensitive to the model performance. The authors define a new payout $v_{GE}(x_j)$ that substitutes the estimated PD with the estimated GE. This is equivalent to the estimated PFI from Eq. (\ref{eq:pfi}). 
\begin{equation*}
v_{GE}(x_j) = \widehat{GE}(\hat{f}, \mathcal{D}_j) - \widehat{GE}(\hat{f}, \mathcal{D}) = \widehat{PFI}_{\hat{f}, j} = v_{PFI}(x_j)
\end{equation*}
We can therefore refer to $v_{GE}(x_j)$ as $v_{PFI}(x_j)$ and regard the SFIMP as an extension to the PFI \cite{casalicchio_featureimp}.

\section{Extending the Framework to Importance Computations}
\label{sec:framework_importance}

\begin{figure}
\centering
 \includegraphics[width = \textwidth]{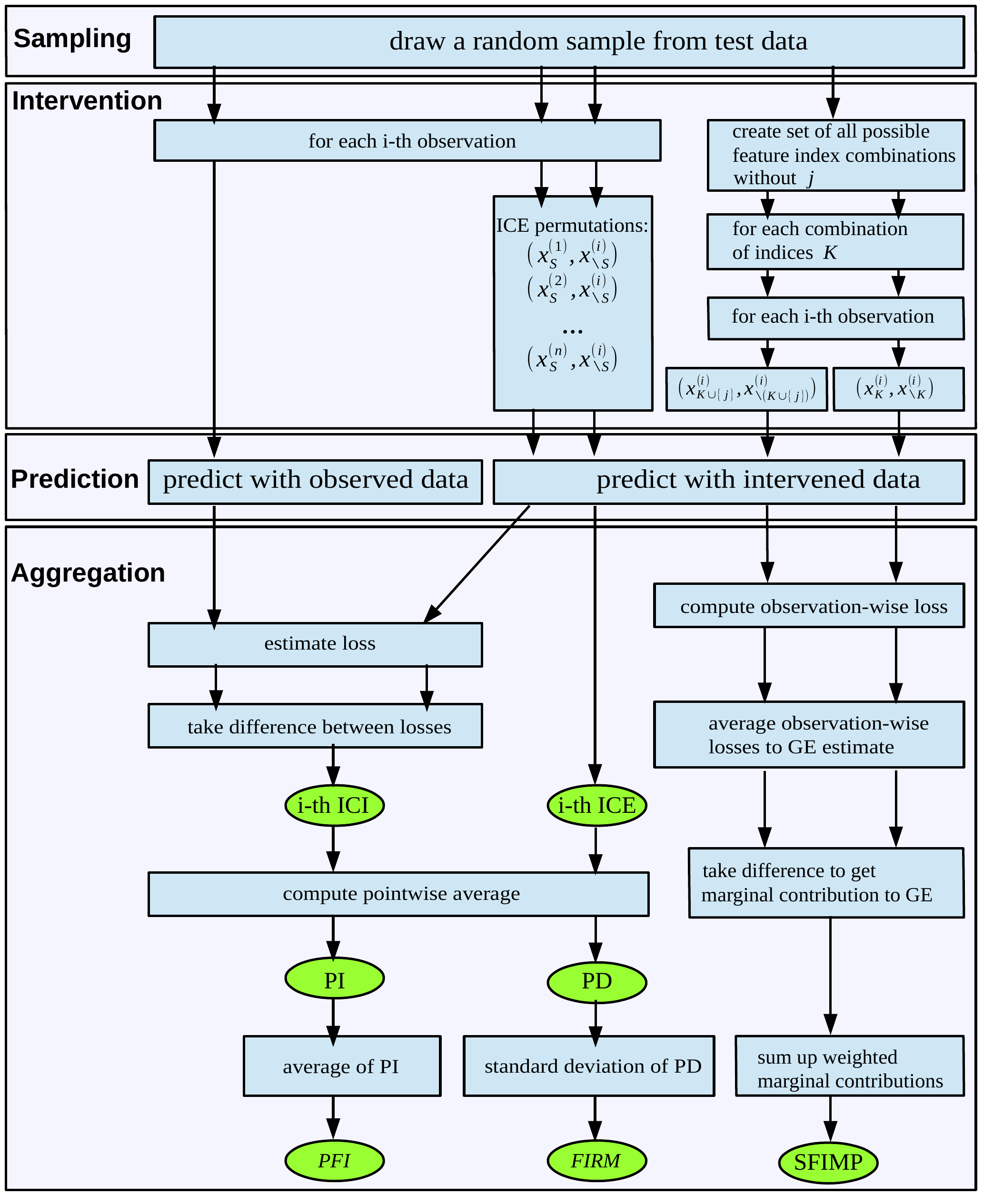}
 \caption{We demonstrate how importance computations are based on the same work stages as effect computations. In the same way as in Fig. 1, we assign the computational steps of all techniques to the corresponding generalized SIPA work stages. Variance-based importance measures such as FIRM measure the variance of a feature effect, i.e., we add a variance computation during the aggregation stage. Performance-based importance measures such as ICI, PI, PFI and SFIMP are based on computing changes in loss after the intervention stage. For reasons of simplicity, we do not differentiate between the actual functions or values and their estimates. \label{framework_importance}}
\end{figure}

Variance-based importance methods measure the variance of feature effect estimates, which we already demonstrated to be based on the SIPA framework. Therefore, we simply add a variance computation during the aggregation stage. Performance-based techniques measure changes in loss, i.e., there are two possible modifications. First, we predict on non-intervened or intervened data (prediction stage). Second, we aggregate predictions to the loss (aggregation stage). In Fig. \ref{framework_importance}, we demonstrate how feature importance computations are based on the same work stages as feature effect computations.

\section{Conclusion}

In recent years, various model-agnostic interpretation methods have been developed. Due to different notations and terminology it is difficult to see how they are related. By deconstructing them into sequential work stages, one discovers striking similarities in their methodologies. We first provided a survey on model-agnostic interpretation methods and then presented the generalized SIPA framework of sequential work stages. First, there is a sampling stage to reduce computational costs. Second, we intervene in the data in order to change the predictions made by the black box model. Third, we predict on intervened or non-intervened data. Fourth, we aggregate the predictions. We embedded multiple methods to estimate the effect (ICE and PD, ALEs, MEs, Shapley values and LIME) and importance (FIRM, PFI, ICI and PI and the SFIMP) of features into the framework. By pointing out how all demonstrated techniques are based on a single methodology, we hope to work towards a more unified view on model-agnostic interpretations and to establish a common ground to discuss them in future work. 

\section*{Acknowledgments}

This work is supported by the Bavarian State Ministry of Science and the
Arts as part of the Centre Digitisation.Bavaria (ZD.B) and
by the German Federal Ministry of Education and Research (BMBF) under
Grant No. 01IS18036A. The authors of this work take full responsibilities for its
content.

\bibliographystyle{splncs04}
\bibliography{../../bibliography/framework_bibfile}

\end{document}